\documentclass[letterpaper, 10 pt, conference]{ieeeconf}

\IEEEoverridecommandlockouts           
 
\usepackage[pdftex]{graphicx}

\usepackage{float} 
\usepackage{amssymb}
\usepackage{amsmath}
\usepackage{bm}
\usepackage{cite}
\usepackage{epsfig}
\usepackage{caption}
\usepackage{color}
\usepackage{psfrag}
\usepackage{cases}
\usepackage{acronym}
\usepackage{setspace}
\usepackage{times}
\usepackage{latexsym}
\usepackage{dblfloatfix}
\usepackage{subcaption}
\usepackage{metalogo}
\usepackage{tabularx}
\usepackage{array}
\usepackage{multirow}
\usepackage{booktabs}
\usepackage{url}

\title{\LARGE \bf
	F-Siamese Tracker: A Frustum-based Double Siamese Network for 3D Single Object Tracking
}
\author{Hao Zou$^{1}$, Jinhao Cui$^{1}$, Xin Kong$^{1}$, Chujuan Zhang$^{1}$, Yong Liu$^{2}$, Feng Wen$^{3}$ and Wanlong Li$^{3}$
	\thanks{$^{1}$Hao Zou, Jinhao Cui, Xin Kong and Chujuan Zhang are with the Institute of Cyber-Systems and Control, Zhejiang University, Zhejiang, 310027, China.}%
	\thanks{$^{2}$Yong Liu is with the State Key Laboratory of Industrial Control Technology and Institute of Cyber-Systems and Control, Zhejiang University, Zhejiang, 310027, China (Yong Liu is the corresponding author, email: yongliu@iipc.zju.edu.cn).}%
	\thanks{$^{3}$Feng Wen and Wanlong Li are with the Huawei Noah's Ark lab}%
}

\begin{document}

\maketitle
\thispagestyle{empty}
\pagestyle{empty}

\begin{abstract}
This paper presents F-Siamese Tracker, a novel approach for single object tracking prominently characterized by more robustly integrating 2D and 3D information to reduce redundant search space. A main challenge in 3D single object tracking is how to reduce search space for generating appropriate 3D candidates. Instead of solely relying on 3D proposals, firstly, our method leverages the Siamese network applied on RGB images to produce 2D region proposals which are then extruded into 3D viewing frustums. Besides, we perform an online accuracy validation on the 3D frustum to generate refined point cloud searching space,  which can be embedded directly into the existing 3D tracking backbone. For efficiency, our approach gains better performance with fewer candidates by reducing search space. In addition,  benefited from introducing the online accuracy validation, for occasional cases with strong occlusions or very sparse points, our approach can still achieve high precision, even when the 2D Siamese tracker loses the target. This approach allows us to set a new state-of-the-art in 3D single object tracking by a significant margin on a sparse outdoor dataset (KITTI tracking). Moreover, experiments on 2D single object tracking show that our framework boosts 2D tracking performance as well.
   
\end{abstract}

\section{Introduction}

Along with the continuous development of autonomous driving, virtual reality and human-computer interaction, single object tracking, as a basic building block in various tasks above, has sparked off public attention in computer vision. For the past few years, many researchers have devoted themselves to studying single object tracking. So far, there are many trackers based on the Siamese network in 2D \cite{c1}, \cite{c2}, \cite{c3} and \cite{c4}, which have obtained desirable performance in the 2D single object tracking. The Siamese network conceives the task of visual object tracking as a general similarity function employing learning through the feature map of both the template branch and the detection branch. In 2D images, convolutional neural networks (CNNs) have fundamentally changed the landscape of computer vision by greatly improving results on many vision tasks such as object detection \cite{c16} \cite{c23}, instance segmentation\cite{c24} and object tracking \cite{c3}. However, since the camera is easily affected by illumination, deformation, occlusions and motion, the occasional cases above do harm to the performance of CNNs and even make invalid.

\begin{figure}[h]
	\centering
	\includegraphics[width = \linewidth]{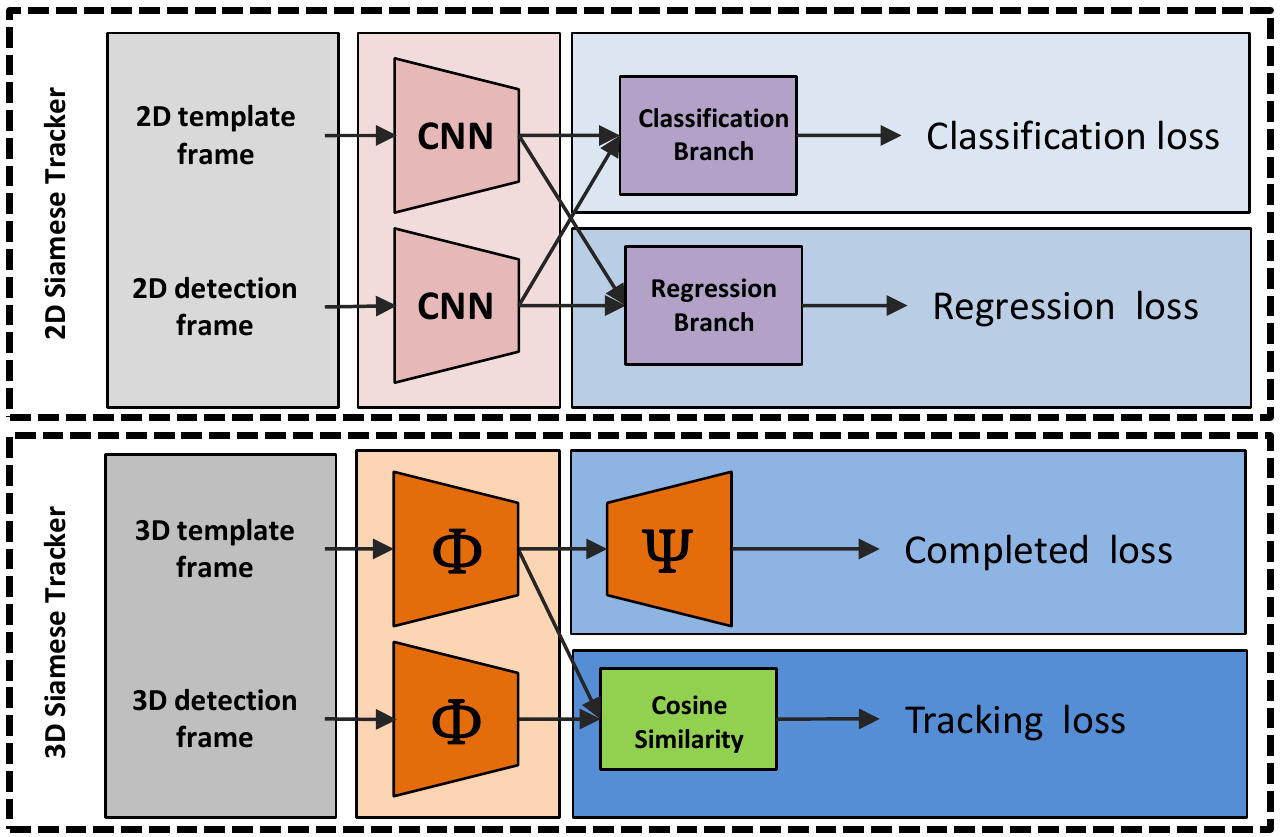}
	\caption{Our proposed a double Siamese network illustration of RGB (top) and point cloud (bottom). In the 2D Siamese tracker, classification score and bounding box regression are obtained via the classification branch and the regression branch, respectively. In the 3D Siamese tracker, the shape completion subnetwork serves as regularization to boost discrimination ability (encoder denoted by $\Phi$ and decoder denoted by $\Psi$). Then we compute the cosine similarity between model shapes and candidate shapes and then generate 3D bounding box.}
	\label{f1}
	
\end{figure}

Inspired by methods above, \cite{c13} takes the lead in coming up with a 3D Siamese network in point clouds. Nevertheless, approaches of this kind carry with them various well-known limitations. The most prominent is that this method, via exhaustive search and lacking RGB information, inevitably has the weakness for the computational complexity in 3D space to generate proposal bounding boxes, which not only results in huge wasting time and space resources but lowers performance. Then \cite{c18} utilizes the 2D Siamese tracker in birds-eye-view (BEV) to generate region proposals in BEV and projects them into the point cloud coordinate for generating candidates. After that, they feed candidates into the 3D Siamese tracker and output the 3D bounding boxes. However, the serial network structure is mostly restricted to relying heavily on 2D tracking results, and BEV loses the fine-grained information in point clouds. We notice that the current autonomous driving systems are mostly equipped with various sensors such as camera and LiDAR. As a consequence, there still requires a proven method of integrating various information for single object tracking.

In this paper, we propose a novel F-Siamese Tracker to address this limitation prominently characterized by fusing RGB and point cloud information.  The proposed method is significant in at least two major respects: reducing redundant search space and solving or relieving the rare case where exist obscured objects and cluttered background  in 2D images as mentioned in \cite{c17}.  To be specific, firstly, we extrude the 2D bounding box from the output by the 2D Siamese tracker into a 3D viewing frustum, then crop this frustum by leveraging the depth value of the 3D template frame. Besides, we perform an online accuracy validation on the frustum to generate refined point cloud searching space,  which can be embedded directly into the existing 3D tracking backbone.

To summarize, the main contributions of this work are listed below in threefold:
\begin{itemize}
	\item We propose a novel end-to-end single object tracking framework taking advantage of various information by more robustly fusing 2D images and 3D point clouds.
	\item We propose an online accuracy validation approach for significantly relieving the dependence on 2D tracking results in the serial network structure and  reducing 3D searching space, which can be fed directly into the existing 3D tracking backbone.
	\item Experiments on the KITTI tracking dataset \cite{c19} show that our method outperforms state-of-the-art methods with remarkable margins, especially for  strong occlusions and very sparse points, thus demonstrating its effectiveness and robustness. Furthermore, experiments on 2D single object tracking show that our framework boosts 2D tracking performance as well.
\end{itemize}

This section will discuss the related work in single object tracking and region proposal methods.
\subsection{Single object tracking}

{\bf 2D-based methods:} Visual object tracking methods have developed rapidly and made great theoretical progress in the past few years, as more datasets have been provided. Public benchmarks like \cite{c5}, \cite{c6}, \cite{c7} provide fair platforms for verifying the effectiveness of visual object tracking approaches. Classic methods based on correlation filtering have achieved remarkable results with the features of strong interpretability and on-the-fly operation \cite{c8}, \cite{c9}. Besides, influenced by the success of deep learning in computer vision,  many end-to-end visual tracking methods have been proposed like \cite{c10}, \cite{c11}. Recently,  \cite{c1} based on a Siamese network proposes a Y-shaped network structure  which joins two network branches: one for the object template and the other for the search region. With its remarkable well-balanced tracking accuracy and efficiency, these methods \cite{c1}, \cite{c2}, \cite{c3}, \cite{c4} have also received  attention in the community. The current state-of-the-art Siamese tracker SiamRPN++ \cite{c3} enhances the tracking performance by presenting a layer-wise feature aggregation structure and depth-wise separable correlation structure, which is one of the pioneering method using deeper CNN such as ResNet-50 \cite{c14}. 
However, this study is limited by the absence of 2D image information and cannot capture geometrical features of the tracked object.

{\bf 3D-based methods}: Compared to 2D trackers, 3D single object tracking methods are still at the primary stage, and relevant work is few. \cite{c15} projects 3D point cloud to BEV, and proposes a deep CNN based on multiple BEV frames to perform various tasks such as detection, tracking and motion forecasting. One major drawback of this approach is that it loses 3D information and causes degradation. Since PointNet \cite{c12} firstly designs an effective learning-based method to directly process the raw point clouds, tracking methods in point clouds are subsequently proposed.  \cite{c13} proposes the first 3D adapted version of the Siamese network for 3D point cloud tracking. They regularize the latent space for a shape completion network \cite{c20}, which leads to the state-of-the-art performance. Nevertheless, approaches of this kind carry with them various well-known limitations. For instance, this method via exhaustive search inevitably has the weakness for extremely high computational complexity in 3D space to generate proposals, which not only results in a huge waste of time and space resources but also lowers performance. Based on SiamRPN \cite{c13}, \cite{c18} proposes an efficient search space using a Region Proposal Network (RPN) in BEV and trains a double Siamese network for tracking. However, BEV loses fine-grained information, making 2D tracking results worse than the ideal, and affecting the final 3D tracking results. Hence, a concise and effective region proposal method is still required to reduce the search space efficiently.

\subsection{Region proposal methods}

In the community, it is commonly noted that the main weakness of two-stage region proposal methods like RCNN \cite{c25} is the paucity of resolving the contradiction of high accuracy but time wasting, due to redundant calculations. In 2D space, in order to reduce the number of proposal regions, Faster-RCNN \cite{c16} proposes RPN, which to some extent relieves the computation expensiveness and redundant storage space in region extraction.  F-PointNet \cite{c17} uses 2D detection result to generate frustums in 3D space, which greatly reduces the search space. However, F-PointNet, with its serial network structure, relies heavily on 2D detection results. \cite{c18} provides an efficient search strategy utilizing the RPN in BEV. However, although they actually leverage additional LiDAR information, they have poor detection for specific  categories like ``Pedestrian'' and ``Cyclist''. The observed result could be attributed to lacking adequate information in two main respects. Firstly, this method does not leverage RGB information. Secondly, objects in these categories above are hardly any points in BEV so as to barely identify. Besides, they rely heavily on 2D tracking results in BEV.
\begin{figure*}[ht]
	\centering
	\includegraphics[width=\linewidth]{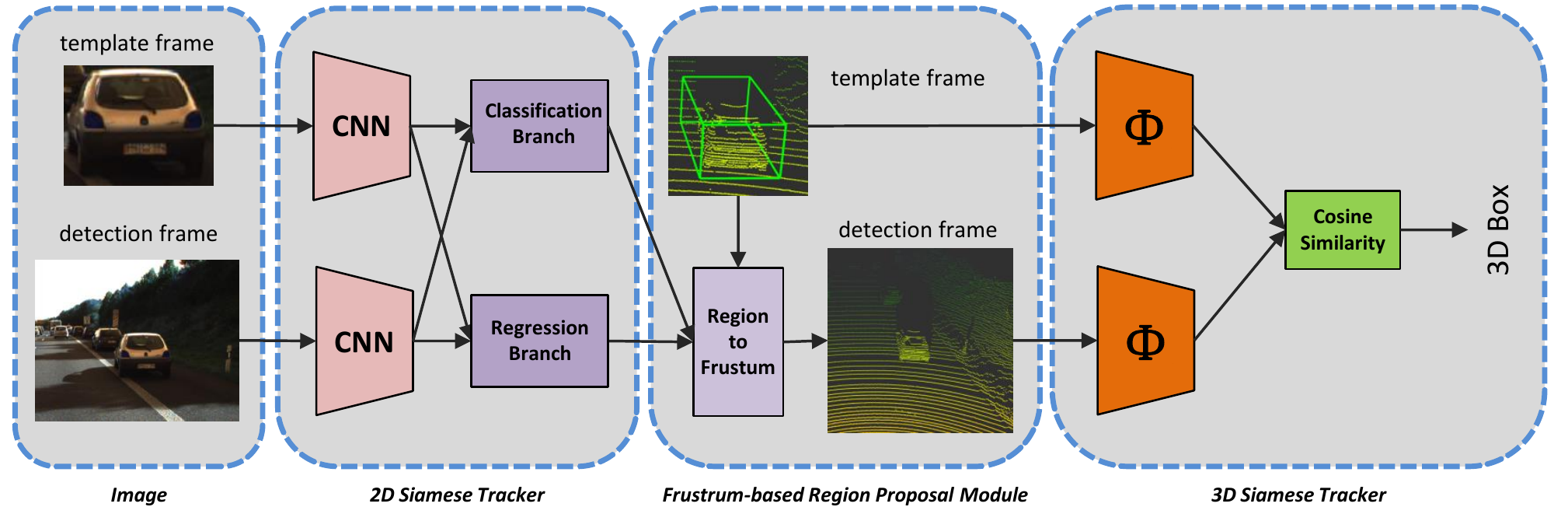}
	\caption{Our F-Siamese Tracker architecture. First, the 2D Siamese Tracker matches the template frame and the detection frame then generates the results of 2D tracking. After that, the Frustum-based Region Proposal Module extrudes these 2D tracking results into 3D viewing frustums and then reduces the volume of the frustum search space via utilizing the depth value of the 3D template frame.
		Finally, the 3D Siamese Tracker serves as encoding point cloud features, then outputs 3D bounding boxes.
	}
	\label{f2}
\end{figure*}

To alleviate the problems above, we propose an approach by making the most of RGB and point cloud information and robustly integrate them. The proposed work takes full advantage of 2D tracking results to reducing search space for the 3D Siamese tracker while avoiding  solely relying on them caused by serial architecture like \cite{c17}.
\section{Methodology}

In this section, considering that the major limitation of 3D single object tracking is lacking appropriate region proposal method and leading to a huge and redundant calculation and time consumption, we propose a novel end-to-end F-Siamese Tracker prominently characterized by fusing RGB and point cloud information. To our best knowledge, our method firstly introduces the Siamese network for integrating RGB and point cloud information in the task of 3D single object tracking.
To be specific, instead of solely relying on 3D proposals, we leverage RGB information to generate the bounding boxes using the mature 2D tracker, then extrude it into a 3D viewing frustum in point cloud coordinate. An overview of our method is shown in Fig. \ref{f1} for training and in Fig. \ref{f2} for inference. Our network architecture (see Fig. \ref{f2}) can be listed as follows: \textit{2D Siamese Tracker}, \textit{Frustum-based Region Proposal Module} and \textit{3D Siamese Tracker}.
\subsection{2D Siamese Tracker}
It is noted that one of top priorities in tracking is how to balance process speed and performance. Hence, the proposed method takes the 2D Siamese tracker for on-the-fly tracking in images. The 2D Siamese tracker, regarding this task as a cross-correlation problem, consists of two parts listed as follows: the siamese feature extraction subnetwork and the region proposal subnetwork. The siamese feature extraction subnetwork includes a fully convolutional network both in the template branch and the detection branch to extract features in the target and search area, respectively. After that, the region proposal subnetwork serves as executing cross-correlation operation between features generated above and then outputs classification and bounding box regression. From all operations above, the 2D Siamese tracker learns a similarity function capable of matching between image in the current frame and target object, then gets the location where target object is in the current frame. 
Advantageously, different 2D Siamese trackers can be flexibly integrated into our framework. Separately, we implement two versions of the tracker in our experiments. One, based on SiamRPN++ \cite{c3} and using ResNet-50 \cite{c14} as backbone, puts emphasis on accuracy. The other, based on SiamRPN \cite{c2} and using AlexNet\cite{c26} as backbone, focuses on the process speed on the contrary.

\begin{figure}[h]
	\centering
	\includegraphics[width = \linewidth]{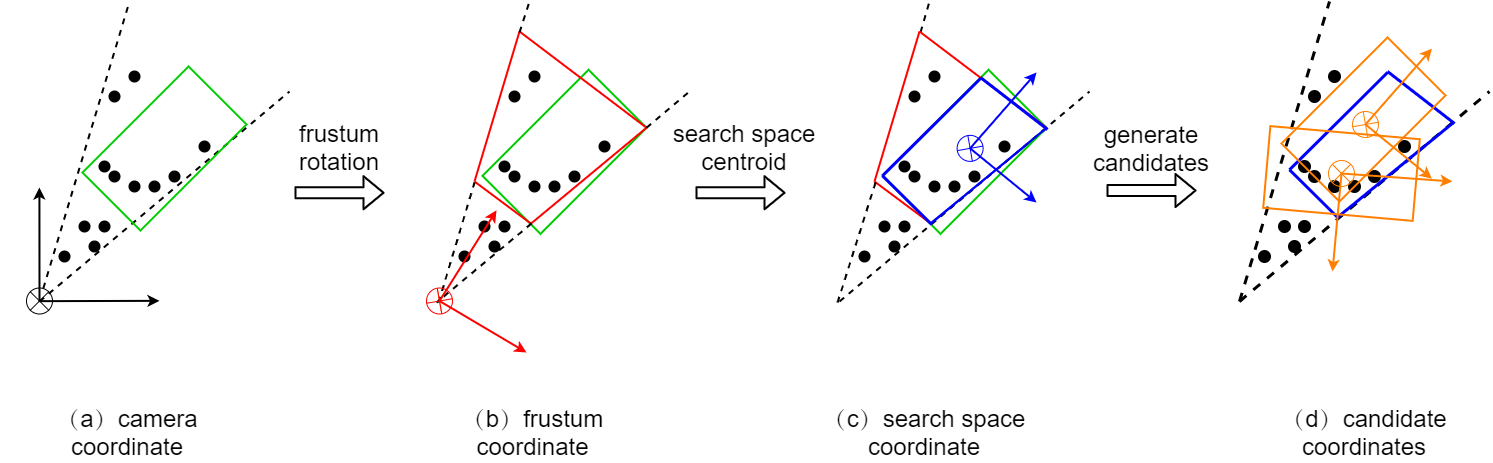}
	\caption{Illustration of the process of producing candidates. Coordinate system is shown listed as follows: (a) default camera coordinate with template box indicated in green; (b) frustum coordinate after rotating the frustum in red to center view; (c) search space coordinate with generated search space shown in blue; (d) candidate coordinates, where orange boxes represent candidates generated in search space.}
	\label{f3}
\end{figure}

\subsection{Frustum-based Region Proposal Module}

After the 2D Siamese Tracker as mentioned above, the Frustum-based Region Proposal Module projects them into point cloud coordinate via camera projection matrix and then extrudes these 2D bounding boxes to 3D viewing frustums. As depicted in Fig. \ref{f3}(a), frustums generated above are vast to the disadvantage of searching.
In view of solid target objects all in continuous and smooth motion, the interval between two frames is limited and the size of target remains constant. Considering that the 3D template frame is continuously updated, our framework uses the previous predicted result as the 3D template frame. As shown in Fig. \ref{f3}(b), our approach can reduce the volume of the frustum search space via utilizing the depth value of the 3D template frame, which not only can solve the occasional case where exist obscured objects and cluttered background in the 2D image as mentioned in \cite{c17}, but also has the capacity of reducing redundant search space, for efficiency. 

\begin{figure*}[ht]
	\centering
	\includegraphics[width=\linewidth]{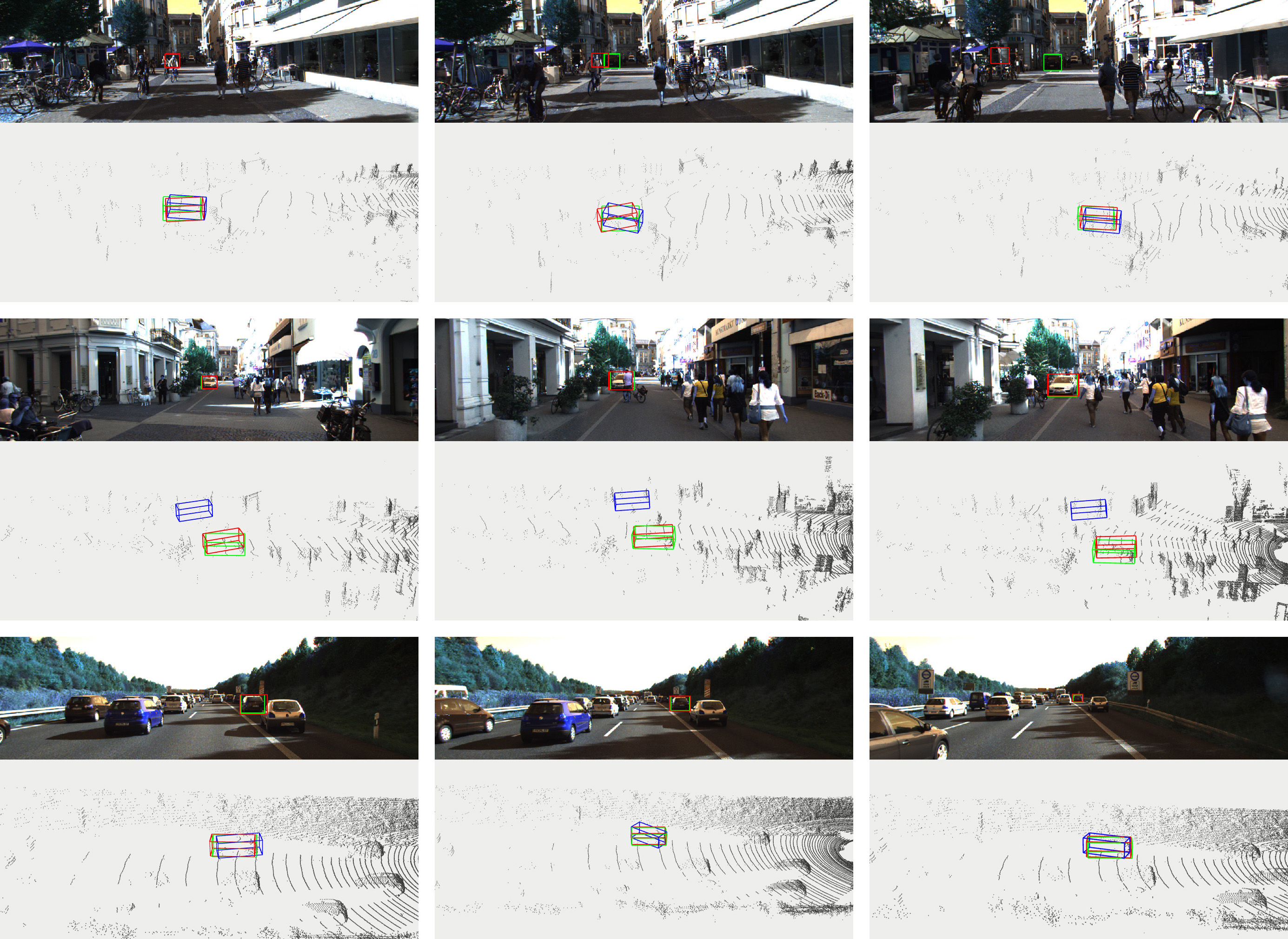}
	\hspace*{2em}\textcolor{green}{---------}  ground truth \hspace*{8em} \textcolor{blue}{---------}  baseline \hspace*{8em}  \textcolor{red}{---------}  our method \\
	\caption{Comparisons of our approach with the state-of-the-art tracker when setting the 3D template frame as the previous predicted result. Experiments show that our method is more robust due to introducing RGB information, and our method can achieve stable tracking even with the very sparse long-range point clouds. Besides, in the occasional case when 2D module passes inaccurate results, our method remains significantly accurate in tracking. }
	\label{f4}
\end{figure*}

However, notwithstanding the satisfied performance of the 2D Siamese tracker, its major limitation  is likely to miss target where there are occasional cases like strong occlusions and illumination variance. In contrast to \cite{c17}, whereas taking generated frustums directly as 3D search space, our approach carries out an online accuracy validation of frustums generated above under the impact of missing target in 2D. As demonstrated in Fig. \ref{f3}(c), the proposed method firstly calculates 3D IoU value (denoted by $\mathcal{V}$) between the intercepted frustum and the 3D template frame. The intersection space of the frustum and the 3D template frame could be utilized when $\mathcal{V}$ is greater than threshold value of 3D IoU  (denoted by $\mathcal{T}$) , otherwise remaining to use the search space in line with \cite{c13}. According the degree of dependency of the 2D Siamese tracker, we adjust the value of $\mathcal{T}$. For instance,  $\mathcal{T}$ equals to 0 shows our method with full dependency of 2D tracking results. On the contrary, our method does not take 2D tracking results into consideration when $\mathcal{T}$ equals to 1. As shown in Fig. \ref{f3}(d), candidates with the same volume of the 3D template frame are exhaustively searched from search space.

To sum up, through steps above, the method in this chapter can significantly avoid or mitigate the weakness of the serial network structure in \cite{c17} and obtain a more streamlined candidates.

\subsection{3D Siamese Tracker}
After Frustum-based Region Proposal Module, we obtain candidates in search space. The points of the interested target are extracted in certain candidate. Fig. \ref{f3}(d) shows that candidate coordinates need to be normalized for translation invariance. Furthermore, the 3D Siamese Tracker takes the normalized point clouds in candidate bounding boxes as input, then outputs the final 3D bounding box.
The 3D Siamese Tracker in our method is consistent with \cite{c13}. \cite{c13} leverages the shape completion network in \cite{c20} as taking raw point clouds as input to realize 3D single object tracking. 

\subsection{Training with Multi-task Losses}
The 2D Siamese Region Proposal Network and the 3D Siamese Tracker are simultaneously trained. After training, the 2D Siamese Region Proposal Network is capable of producing 2D region proposals quickly and accurately. Then we feed them into the 3D Siamese Tracker to compare and select the best candidate. Our network architecture adopts the method of multi-task losses to optimize the whole network. The loss function could be formulated as 
\begin{equation}
L o s s=L_{2 d}+L_{3 d}
\end{equation}
\begin{equation}
L_{2 d}=\lambda_{c l f} \cdot L_{c l f}+\lambda_{r e g} \cdot L_{r e g}
\end{equation}
\begin{equation}
L_{3 d}=\lambda_{t r} \cdot L_{t r}+\lambda_{c o m p} \cdot L_{c o m p}
\label{formula4}
\end{equation}
where $L_{c l f}$ is  the cross-entropy loss for classification, $L_{r e g}$ is the smooth L1 loss for regression, $ L_{t r}$ is the MSE loss for tracking and $ L_{c o m p}$ is the L2 loss for shape completion. During training, the target is to minimize the loss using the Adam optimizer \cite{c21} with the initial learning rate of $10^{-4}$, $\beta1$ of 0.9 and the batch size of $32$. $\lambda_{c l f}, \lambda_{r e g},  \lambda_{t r} , \lambda_{c o m p}$ equal to $1, 1.2, 1, 10^{-6}$ respectively.
\begin{table*}[ht]
	\centering
	\setlength{\tabcolsep}{4mm}{
		\begin{tabular}{c||cccccc}
			\hline
			\multirow{3}{*}{Method}      & \multicolumn{6}{c}{Class} \\ \cline{2-7} 
			& \multicolumn{2}{c}{Car}                & \multicolumn{2}{c}{Pedestrian}          & \multicolumn{2}{c}{Cyclist}             \\
			& \multicolumn{1}{l}{Success} & Precision & \multicolumn{1}{l}{Success} & Precision & \multicolumn{1}{l}{Success} & Precision \\ \hline
			Origin 3D Siamese Tracker + GT & 78.46   &  82.96  & -  &  - & -  & -  \\
			Origin 3D Siamese Tracker + PR & 24.66 &     30.67      &  -                          &      -     & -                 &     -      \\ \hline
			Ours + GT          & \textbf{81.58}   &  \textbf{87.32}  & \textbf{61.85}  & \textbf{70.36} & \textbf{88.66}  & \textbf{99.67}  \\
			Ours + PR         & \textbf{37.12}   &  \textbf{50.60}  & \textbf{16.28} &  \textbf{32.28} & \textbf{47.03}  & \textbf{77.26}  \\ \hline
			
	\end{tabular}}
	\caption{Comparisons of the performance of 3D single object tracking between our method and state-of-the-art. + GT denotes adopting the current ground truth as the 3D template frame. + PR denotes adopting the previous predicted result as the 3D template frame.}
	\label{t1}
\end{table*}

\begin{table}[ht]
	\centering
	\setlength{\tabcolsep}{5mm}{
		\begin{tabular}{c||clclcl}
			\hline
			\multirow{3}{*}{Method}                & \multicolumn{2}{c}{Class}              \\ \cline{2-3} 
			& \multicolumn{2}{c}{Car}                \\
			& \multicolumn{1}{l}{Success} & Precision \\ \hline
			SiamRPN\cite{c2}           & 63.80         &  70.00        \\
			
			SiamRPN++\cite{c3}           & 64.12        &  71.35         \\
			
			Our & \textbf{79.42}                  &  \textbf{85.24}         \\ \hline\end{tabular}}
	\caption{Comparisons of the performance of 2D single object tracking between our model and \cite{c2}, \cite{c3} by projecting the generated 3D bounding box to image coordinate to obtain 2D bounding box.}
	\label{t2}
\end{table}


\section{Experiments}
In the section that follows, we evaluate our approach by comparing with the current state-of-the-art method \cite{c13}. 
The main outcome to emerge from our experiments is that our model improves the performance of 3D single object tracking via an effective approach for reducing search space.

\subsection{Implementation Details}
\textbf{Dataset:}
Here, we evaluate the proposed work on the KITTI tracking dataset \cite{c19}. Following \cite{c13}, this dataset is divided into these three parts: 0-16 for training, 17-18 for validation  and 19-20 for testing. We use these categories: `Car', `Pedestrian' and `Cyclist' and then combine all the scenes located the tracking target object into a tracklet. 

\textbf{Evaluation Metric:}
Following previous works \cite{c13}, we use One Pass Evaluation (OPE) \cite{c22} as the metric for evaluation. It defines the overlap as the IoU of a bounding box with its ground truth, and the error as the distance between both centers. The Success and the Precision metrics are defined using the overlap and error Area Under Curve (AUC).
\subsection{Quantitative and Qualitative Results}
Table. \ref{t1} reports an overview of the performance of our architecture compared to the origin 3D Siamese tracker \cite{c13} using two different 3D template frames: one is the current ground truth and the other is the previous predicted result. The output of our network is visualized in Fig. \ref{f4}. From Fig. \ref{f4} we can see that 3D object tracking might be under very challenging cases,  such as the very sparse point cloud, obstacled object and invalid 2D tracker.

We choose SiamRPN++ as the 2D tracker, and the threshold value $\mathcal{T}$ of 3D IoU should be set. When 3D IoU between the generated frustum and the 3D template frame is greater than $\mathcal{T}$, 3D search space is reduced to the intersection space, and our approach generates $\mathcal{N}$ candidates in the 3D detection frame, otherwise search space stays constant and our approach generates 147 3D candidates in line with \cite{c13}. In the testing stage, however, the origin 3D Siamese Tracker \cite{c13} takes the current ground truth as the 3D template frame, instead of the previous predicted result. Consequently, we change the 3D template frame to the previous predicted result and evaluate the performance of \cite{c13}. Our experiments set $\mathcal{T}$ to 0.8 for using current ground truth as the 3D template frame, while setting $\mathcal{T}$  to 0.2 for using previous predicted result. 
In the proposed method, we set $\mathcal{N}$ to 72 far less than that in baseline.
\\ \hspace*{1em} What stands out in Table. \ref{t1} is that the proposed method performs better than state-of-the-art for all settings in our experiments.
Specifically, our method obtains 50.6\% precision, which outperforms precision 30.6\% of baseline by nearly 20\% when using previous predicted result as the 3D template frame. 
We also test 2D single object tracking by projecting the results in 3D space into images at the same time. Following settings in line with \cite{c13}, Table. \ref{t2} reports that our method outperforms than 2D single object tracking state-of-the-art \cite{c3} as well. Our method achieves better performance, and increases the success rate to 80.42\% and the precision rate to 85.24\% in the category of car.

Taken together, this remarkable improvement of precision both in 2D and 3D proves that the robustness and accuracy of the proposed method.

\subsection{Ablation Studies}
In this subsection, we conduct extensive ablation experiments to analyze the performance of the proposed method for introducing the image information into the 3D single object tracking. 

\textbf{Threshold of 3D IoU:} To begin with, we follow the standard-settings provided by \cite{c13}, and conduct an ablation study to analyze the effects of inverse thresholds $\mathcal{T}$ of 3D IoU. Fig. \ref{level.sub.1} and Fig. \ref{level.sub.3} illustrates the performance by a large margin among different $\mathcal{T}$. When using the previous predicted result as the 3D template frame, setting $\mathcal{T}$ to 0.1 tends to have the best performance in our experiments. A possible explanation for this might be that baseline performs not very well when using the previous predicted result rather than ground truth. Hence, introducing RGB information seems to significantly improve the results. Besides, when using current ground truth as reference,  setting $\mathcal{T}$ to 0.8 tends to have the best performance in our experiments. This result is likely to be related to that the performance of baseline is probably good enough, introducing RGB information has limited performance improvement. \\ \hspace*{1em}

\begin{table}[]
	\centering
	\setlength{\tabcolsep}{2mm}{
		\begin{tabular}{c|cc}
			\hline
			\multirow{3}{*}{Method}         & \multicolumn{2}{c}{Class} \\ \cline{2-3} 
			& \multicolumn{2}{c}{Car}   \\
			& Success     & Precision    \\ \hline
			Our + 27                        & 22.79       & 30.61        \\
			Our + 32                        & 25.54       & 34.21        \\
			Our + 50                        & 28.79       & 38.58        \\ \hline
			Origin 3D Siamese tracker + 147 & 24.66       & 30.67        \\ \hline
	\end{tabular}}
	\caption{Comparisons of the performance of 3D single object tracking between our model and state-of-the-art with different quantity of candidates.  + $\mathcal{N}$ denotes setting $\mathcal{N}$ candidates. }
	\label{t3}
\end{table}

\textbf{Quantity of Candidates: }Furthermore, we also study the effects of the inverse quantity of candidates  $\mathcal{N}$, considering the baseline lacking an effective region proposal method, we set $\mathcal{T}$ to 0.2 when using the previous predicted result as reference, and to 0.8 when using ground truth as reference. Fig. \ref{level.sub.2} and Fig. \ref{level.sub.4} show that there is the best performance when $\mathcal{N}$ equals to 72, and more candidates have little effect on the improvement of the performance.

Taking into account the efficiency problems in practical application, we conduct an ablation study on the number of candidates. We adopt the previous predicted result as the 3D template frame. We replace SiamRPN++ \cite{c3} with SiamRPN \cite{c2} as the 2D Siamese tracker and set $\mathcal{T}$ equals to 0. Table. \ref{t3} presents that our approach significantly improves efficiency with less candidates. Specifically, when setting $\mathcal{N}$ to 32, our method with higher precision is nearly twice fast than baseline. In our experiments on GTX 1080Ti GPU, the operation time of our method in 1000 frames is 3.37 minutes, less than 7.45 minutes of baseline.

\section{Conclusion}
 This paper has presented a unified framework named F-Siamese Tracker to train an end-to-end deep Siamese network for 3D tracking. Via robustly integrating RGB and point cloud information, the search space of the 3D Siamese tracker is significantly reduced by introducing a mature 2D single object tracking approach, which greatly improves the performance of 3D tracking. Extensive experiments with  state-of-the-art performance on KITTI tracking dataset demonstrate the effectiveness and generality of our approach. Further research might explore how to further integrate RGB and point cloud information into the Siamese network. We believe the proposed framework can, in principle, advance the research of 3D single object tracking in the community. 
 
\section{Acknowledgement}
This work is supported by the National Natural Science Foundation of China under Grant 61836015.

\addtolength{\topmargin}{0.249cm}
\bibliographystyle{ieeetr}
\bibliography{egbib}

\end{document}